\documentclass[a4paper,fleqn]{cas-dc}

\usepackage[authoryear,longnamesfirst]{natbib}
\usepackage{times}
\usepackage{epsfig}
\usepackage{graphicx}
\usepackage{amsmath}
\usepackage{amssymb}
\usepackage[ruled,linesnumbered]{algorithm2e}
\usepackage{bm}
\usepackage{booktabs}
\usepackage[switch]{lineno}
\usepackage[switch]{lineno}

\usepackage{microtype}
\usepackage{graphicx}
\usepackage{subfigure}
\usepackage{multirow} 
\usepackage{balance}
\graphicspath{{./image/}}
\def\tsc#1{\csdef{#1}{\textsc{\lowercase{#1}}\xspace}}
\tsc{WGM}
\tsc{QE}

\begin{document}
\let\WriteBookmarks\relax
\def\floatpagepagefraction{1}
\def\textpagefraction{.001}

\shorttitle{MPSN: Motion-aware Pseudo-Siamese Network for Indoor Video Head Detection in Buildings}    

\shortauthors{Kailai Sun et al.}

\title[mode=title]{MPSN: Motion-aware Pseudo-Siamese Network for Indoor Video Head Detection in Buildings}  



%

\author[1]{Kailai Sun}[type=editor,style=chinese,orcid=0000-0003-1648-3409]


\fnmark[1]

\ead{skl18@mails.tsinghua.edu.cn}

\author[1]{Xiaoteng Ma}[type=editor,style=chinese,orcid=0000-0002-7250-6268]

\fnmark[1]

\ead{ma-xt17@mails.tsinghua.edu.cn}


\author[1]{Peng Liu}[type=editor,style=chinese]
\fnmark[1]
\ead{liup20@mails.tsinghua.edu.cn}

\author[1]{Qianchuan Zhao}[type=editor,style=chinese,orcid=0000-0002-7952-5621]
\cormark[1]


\ead{zhaoqc@mail.tsinghua.edu.cn}



\affiliation[1]{organization={Center for Intelligent and Networked Systems, Department of Automation, BNRist, Tsinghua University},
            city={Beijing},
            postcode={100084}, 
            country={China}}

\cortext[1]{Corresponding author.}

\fntext[1]{These authors contributed equally to this work.}


\begin{abstract}
Indoor video head detection is an essential component of building occupancy detection. While deep \textcolor{black}{learning} models have achieved remarkable progress in general object detection, \textcolor{black}{their performance is limited in complex indoor scenes}. Indoor surveillance video\textcolor{black}{s often contain} cluttered background objects, among which heads have small scales and diverse poses. In this paper, we propose \textbf{M}otion-aware \textbf{P}seudo-\textbf{S}iamese \textbf{N}etwork (MPSN), an end-to-end approach that leverages head motion information to guide deep \textcolor{black}{learning} models to extract effective head features in indoor scenarios. By taking the pixel-wise difference \textcolor{black}{between} adjacent frames as the auxiliary input, MPSN effectively enhances human head motion information and removes irrelevant \textcolor{black}{background} objects. \textcolor{black}{Its performance is compared with that of existing methods.} Compared with \textcolor{black}{existing} methods, \textcolor{black}{MPSN} achieves superior performance on two indoor video datasets. \textcolor{black}{It} successfully suppresses static background objects and highlights the moving instances, especially human heads in indoor videos. Different methods to capture head motion also were compared. MPSN demonstrates simplicity and flexibility \textcolor{black}{compared to other methods}. \textcolor{black}{We validate its robustness through adversarial experiments} with a mathematical solution of small perturbations for robust model selection. \textcolor{black}{To} confirm its potential in building control systems, we apply MPSN to occupancy counting. \textcolor{black}{The} code is  available  at \emph{https://github.com/pl-share/MPSN}.
\end{abstract}


\maketitle

\section{INTRODUCTION}
\label{section1}

The human dimension information plays a significant role in efficient building energy-saving and comfortable indoor environments \citep{RUEDA2020106966,SUN2020109965}. Recent studies showed that control strategies based on occupancy information can save building energy by approximately 20$-$45$\%$, and improve thermal comfort by 29.1$\%$ \citep{PANG2020115727}. Occupancy information ensures a closed-loop feedback strategy \citep{RN475} to control building heating, ventilation, and air-conditioning (HVAC) and lighting systems.

Vision-based occupancy detection to obtain accurate and robust occupancy information in buildings \textcolor{black}{has recently become a hot research topic} \citep{SUN2020109965,choi2021review}. Vision-based methods usually capture images/videos \textcolor{black}{using} cameras, and then apply image processing, video analysis, and deep learning technologies to sense people in buildings. These methods are mainly divided into three categories: body, face, and head detection. Head detection has been the perceived center point of people detection in complex indoor scenes. The limitations of body and face detection methods have gradually been exposed. Specifically, body detection methods extract the body features for recognition, which suffers from inter-class and intra-class occlusions \citep{RN609,choi2021application}. Face detection methods rely on facial features \textcolor{black}{; these mothods} fail when the person to be detected is \textcolor{black}{not facing the camera}. Compared with the former two methods, head detection has a wider range of applications because human heads are visible and reliable in complex indoor scenes. Indoor occupancy detection methods in buildings \citep{trivedi2020occupancy,RN475,SUN2022111593} focus on the head part instead of the whole human body on account of the severe occlusion.

\begin{figure}[t]
  \centering
  \includegraphics[scale=0.24]{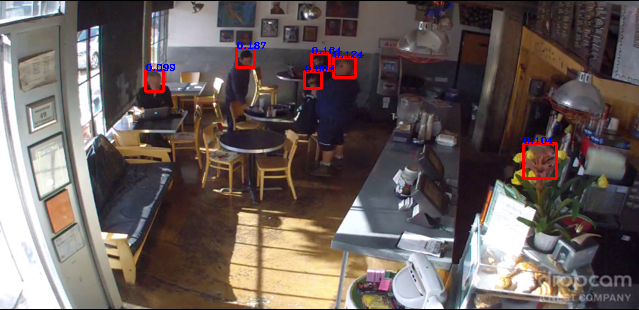}
  \includegraphics[scale=0.24]{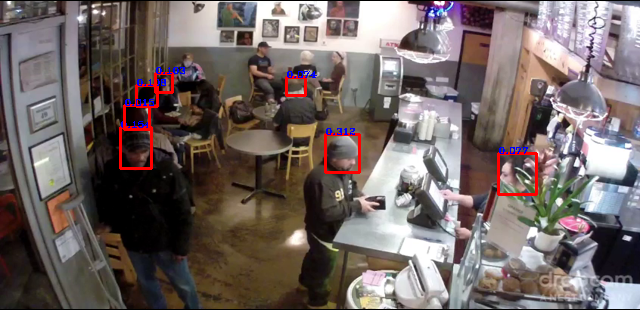}

  FCHD (Baseline)
  \vspace{0.7em}
  
  \includegraphics[scale=0.18]{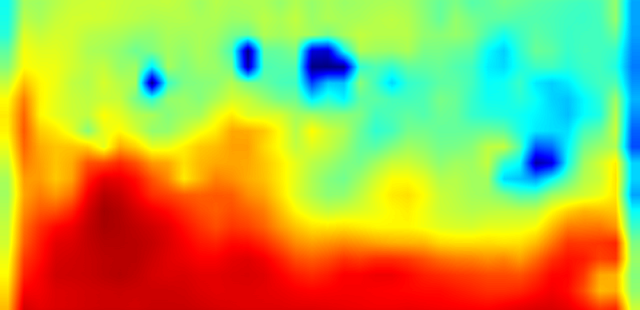}
  \includegraphics[scale=0.18]{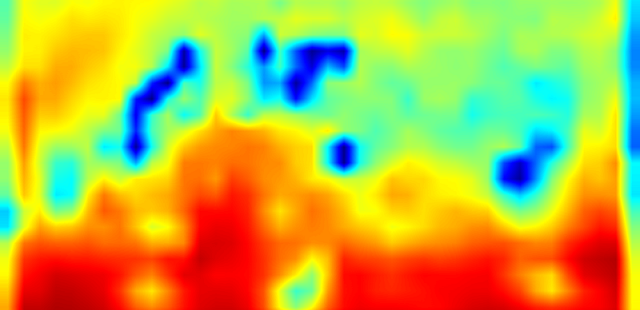}
  CAM of FCHD
  \vspace{0.5em}
  
  \includegraphics[scale=0.18]{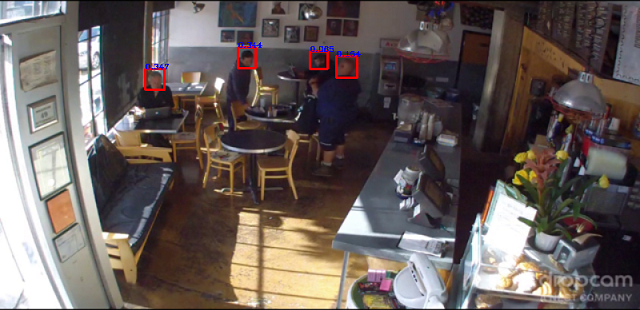}
  \includegraphics[scale=0.18]{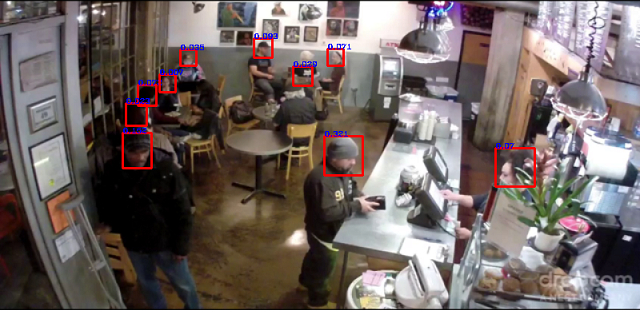}

  MPSN (Ours)
  \vspace{0.5em}

  \includegraphics[scale=0.18]{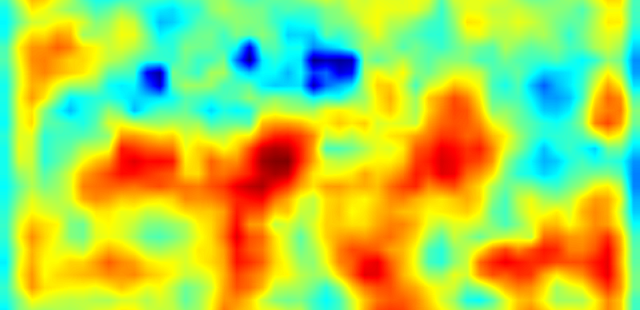}
  \includegraphics[scale=0.18]{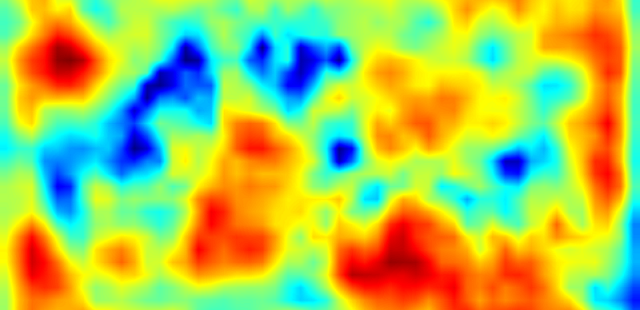}
  CAM of MPSN

  \vspace{0.5em}
  
\includegraphics[scale=0.3]{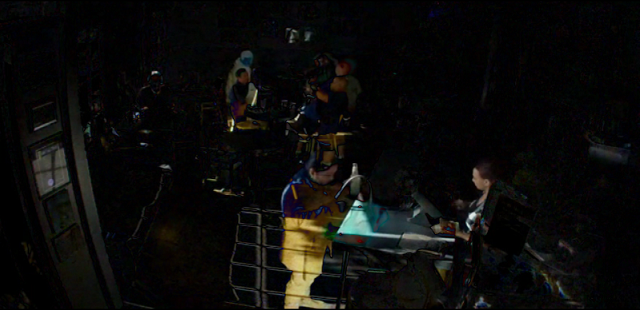}
  \includegraphics[scale=0.3]{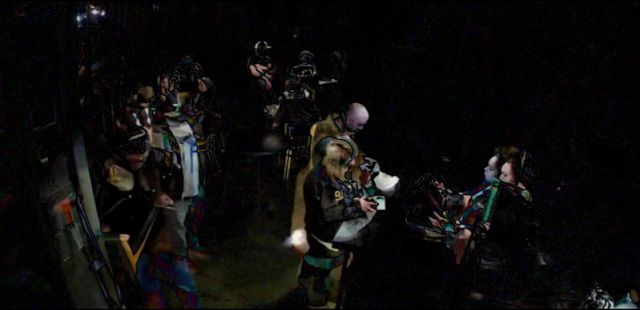}

  Frame Difference
  \vspace{0.5em}

  \label{show1}
  \caption{Comparison of MPSN with the baseline method FCHD \citep{DBLP:journals/corr/abs-1809-08766}. Two columns show two respective instances in the crowd Brainwash dataset~\citep{DBLP:journals/corr/StewartA15}. In the first column, when similar objects (flowers) exist, MPSN can suppress the static regions and filter many false positives. In the second column, the single image detector fails to propose accurate boxes when heads are small. Our MPSN can enhance the motion regions and reduce many false negatives. For interpretation purposes, we visualize the features (heatmaps) of the last CNN layer, which are generated by the class activation map (CAM) method \citep{DBLP:journals/corr/ZhouKLOT15}. Deep blue regions of the heatmaps represent the locations of heads in the feature space. 
}
  \vspace{-2em}
\end{figure}

Although many head detectors \textcolor{black}{have made} great progress in general scenes \citep{ISI:000380414100323,DBLP:journals/corr/abs-1809-03336,liu2021head}, the head detection task is still challenging in indoor scenes. As shown in Fig.~\ref{show1}, many background objects (e.g., black bags, balls, flowers) have head-like sizes, colors, and textures, resulting in a high false-positive rate (FPR). On the other hand, due to the relatively small scales, it is hard to detect heads with a high confidence score in crowded scenes. Moreover, moving head\textcolor{black}{s} cause significant variations in scale, pose, texture, and illumination, which increases the false-negative rate (FNR). \textcolor{black}{Research} on multi-scale features \citep{liu2021samnet} and attention mechanisms \citep{woo2018cbam,shen2019indoor} \textcolor{black}{aims} to handle the multi-scale objects and similar background objects. However, existing studies mainly focus on head detection in static images \citep{DBLP:journals/corr/abs-1809-08766,ISI:000521828604112,DBLP:journals/corr/StewartA15}, which \textcolor{black}{is inadequate} in indoor video scenarios. 

\textcolor{black}{Many video head detection approaches \citep{ISI:000604225500001,RN475,Sundararaman_2021_CVPR} aim to refine the final results by designing trackers and hand-crafted box-level rules between neighboring frames \citep{8237592}. But these box-level methods only rely on single detectors and are not optimized jointly \citep{10ccc}. Box-level methods would be inaccurate when the appearance of objects dramatically changes, especially as objects are occluded, or the interval between two nearby frames is large.}

\textcolor{black}{We notice that human heads are often in motion while the background objects are almost always static. To solve these issues, the following question needs to be answered:  \emph{can human head motion be utilized to enhance the features while suppressing background information?}} This study gives us a confirmatory answer. The head motion information is captured by the \textcolor{black}{pixel-wise difference between} adjacent frames. The difference is sensitive to moving objects even though they are small in complex scenes. Besides, the motion information \textcolor{black}{can be used} to filter static objects. If static instances \textcolor{black}{are suppressed}  while enhancing moving instances, the FPR and FNR \textcolor{black}{can} be effectively reduced.

We propose \textbf{M}otion-aware \textbf{P}seudo-\textbf{S}iamese \textbf{N}etwork (MPSN) for indoor video head detection, an accurate and flexible framework for video head detection \textcolor{black}{at the pixel level}. It can robustly detect head regions in a sequence of frames, even when the time intervals of frames are large. First, we extract frame difference features between neighboring frames, which are sensitive to motion regions (see Fig.~\ref{show1} for examples in the real dataset). Second, we propose a pseudo-siamese network to extract similar features between frame difference and the original image. Third, these features are aggregated with a constraint \textcolor{black}{that} can suppress false positives and recover false negatives. Finally, detection methods, such as region proposal network (RPN) \citep{DBLP:journals/corr/RenHG015}, are cascaded to generate proposal head boxes.

\textcolor{black}{Besides, we utilize the pixel-level motion feature to guide the single-frame detector to propose robust head boxes. Unlike box-level methods (post-processing), our approach works at feature and pixel levels (pre-processing). Previous head detection approaches at feature or pixel levels are rare.} To our best knowledge, MPSN is the first to jointly train the current frame and \textcolor{black}{the pixel-level} motion information into an end-to-end CNN network in head detection. This \textcolor{black}{study} aims to solve the indoor video head detection \textcolor{black}{problems}, such as similar static objects and diverse head samples. The contributions of this \textcolor{black}{study are as follows}: 1) We propose a pixel-level motion-aware pseudo-siamese network to learn the more robust head features. 2) With more effective motion features, we demonstrate MPSN on the crowd Brainwash dataset \citep{DBLP:journals/corr/StewartA15} and the Restaurant dataset \citep{9150687} with superior performance and low memory requirements. 3) As a general head detection framework, we demonstrate the flexibility of MPSN with different backbone networks (VGGNet16 \citep{vgg} , MobileNetv2 \citep{DBLP:journals/corr/abs-1801-04381}, Resnet18 \citep{DBLP:journals/corr/HeZRS15}). 4) To evaluate the robustness of MPSN, we implement adversarial samples to MPSN, and provide the mathematical solution of small perturbations to select more robust models. 5) We apply MPSN to occupancy counting, which confirms its potential in practical building control systems.

\section{RELATED WORK}

\subsection{Generic \textcolor{black}{object detection}}
In recent years, deep learning technologies (e.g., CNN and transformer) have achieved state-of-the-art performances in object detection tasks. These \textcolor{black}{CNN-based} methods can be divided into two categories: two-stage approach and one-stage approach. The two-stage methods, such as RCNN \citep{TIEN2022118336} and R-FCN \citep{DBLP:journals/corr/DaiLHS16}, usually propose enough ROI (region of interest) boxes that contain potential objects, predict them as specific categories, and refine the offsets and scales of bounding boxes. The one-stage methods, such as YOLO \citep{choi2021application}, SSD \citep{Zhang_2018_CVPR}, and CenterNet \citep{DBLP:journals/corr/abs-1904-08189}, usually jointly predict the probability scores and regress bounding boxes together. The two-stage methods can achieve higher accuracy, while the one-stage methods have less computational complexity. Recently, DETR \citep{dai2021dynamic} directly \textcolor{black}{utilized} transformer decoder and bipartite matching strategies to predict the joint class, position, and bounding boxes. However, transformer networks require a large number of training parameters and samples, which hinders \textcolor{black}{their deployment in} resource-constrained environment\textcolor{black}{s}~\citep{khan2021transformers}.

\subsection{Indoor \textcolor{black}{occupancy detection}}
The applications of indoor occupancy detection methods include building energy saving, indoor environment comfort, and security management.
Indoor occupancy detection is a challenging task involving people detection, video analysis, etc. Most Studies \citep{Huang2020DevelopmentOC,MUTIS2020103237,DBLP:journals/corr/abs-1809-08766,RN519,choi2021application} used general object detectors (YOLO, Faster RCNN, etc.) to detect indoor occupancy. 

However, the indoor scenes are extremely complex \textcolor{black}{owing to human bodies being} occluded by desks, chairs, and other people \citep{SUN2022111593,choi2021application}. Therefore, to avoid the severe occlusion problem, occupancy detection methods based on head detection have been developed \citep{DBLP:journals/corr/abs-1809-08766,RN475,GUAN2015181,acquaah2020occupancy}. 
Object/head detection methods can be cascaded with video analysis technologies to obtain higher detection performance \citep{SUN2022111593,dino2019video}. 

\subsection{Indoor \textcolor{black}{head detection}}
Most related studies treat indoor head detection as a subtask of object detection. Early methods \textcolor{black}{used} hand-crafted features (Haar features \citep{INSPEC:7176899}, aggregate channel features \citep{RN610}, histogram of oriented gradients features\citep{RN475}, etc.) and classifiers to detect human heads. With the rapid development of deep learning, the mainstream of head detection method \textcolor{black}{switched to} CNN-based object detection methods \citep{DBLP:journals/corr/StewartA15,chouai2021new,khan2020robust,ISI:000604225500001,babu2017switching}. 

In practice, different from generic objects, human heads have special properties. Thus, the specific anchor size selection strategy~\citep{DBLP:journals/corr/abs-1809-08766,Khan2021}, multi-scale features aggregation strategy \citep{ISI:000380414100323,xiang2017joint}, attention selection strategy \citep{li2019dsfd,shen2019indoor}, and the relationship between heads and people matching strategy \citep{RN609} can effectively improve detection performance.

\section{METHODS}\label{Method}

\begin{figure*}
\begin{center}
\includegraphics[width=1\linewidth]{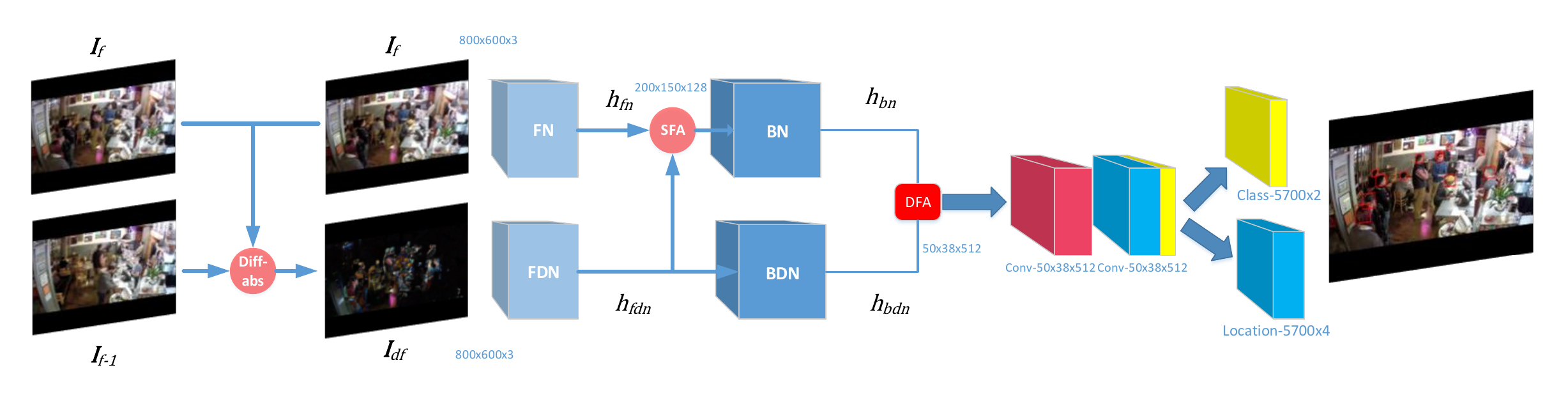}
\end{center}
\vspace{-2em}
\caption{Framework of MPSN. Firstly, the original frame $\bm{I_{f}}$ and the frame difference $\bm{I_{df}}$ are \textcolor{black}{fed in parallel}  to the pseudo-siamese network. The paired networks (FN and FDN, BN and BDN) have the same architectures but different weights. Secondly, the features are aggregated with our SFA and DFA at different pixel levels. Finally, the boxes and classes are predicted simultaneously with RPN.}
\label{fig1}
\vspace{-1em}
\end{figure*}

Given an image, we have a feature extraction network $\emph{N}_{feat}$, and a detection network $\emph{N}_{det}$. The output for input image $\bm{I}$ is $\emph{N}_{det}(\bm{h})$, where $\bm{h}=\emph{N}_{feat}(\bm{I})$. The key \textcolor{black}{concept of this study} is to apply motion information to suppress static background objects and focus on the head instances \textcolor{black}{at} the pixel level.

The inference pipeline is depicted in Fig.~\ref{fig1}, consisting of four modules: 
1) The \emph{motion estimation} module is developed to estimate the head motion information, which is fed to the pseudo-siamese network. 2) The \emph{feature aggregation} module is proposed to extract robust head motion features. 3) The \emph{gradient backpropagation} module which is presented in Appendix can explain the gradient principle of MPSN. 4) Finally, The \emph{detection network} module decodes the aggregated features to head boxes.
 

\subsection{Motion estimation} \label{Motion}

As mentioned in Section~\ref{section1}, the human head would not be static for a long time. We introduce prior motion knowledge to guide the network $\emph{N}_{feat}$ to extract effective head features. For further applications, we use simple background subtraction methods to roughly extract motion areas. We define the frame difference operator as 

\begin{equation}\label{motion}
\bm{I_{df}}=|\bm{I_f}-\bm{I_{f-1}}|,\bm{f}=1,\dots ,N, 
\end{equation}
where $\bm{I_f}$ is the $\bm{f}{\rm th}$ frame in an image sequence, and $N$ is the sequence length. \textcolor{black}{For color images, we conduct the same operation on three channels (RGB channels) by using Eq.~\eqref{motion}, respectively}. Although the time intervals between $\bm{I_{f+1}}$ and $\bm{I_f}$ have large variations in applications, this operator is sensitive to motion pixels and contains spatial location information. 

Optical flow has been widely applied in video analysis. It helps to estimate video motion information. In this paper, we define the motion field by the optical flow estimation algorithm $\mathcal{F}$ \citep{teed2020raft,ilg2017flownet,lucas1981iterative} as
\begin{equation}
\bm{I_{df}}=\mathcal{F}(\bm{I_f},\bm{I_{f-1}}),\bm{f}=1,\dots ,N,
\end{equation}

The rough motion features will be further extracted by the following pseudo-siamese network. 

\subsection{Feature aggregation}
The original frame $\bm{I_{f}}$ and frame difference image $\bm{I_{df}}$ are \textcolor{black}{fed in parallel} to the feature extraction network $\emph{N}_{feat}$. In other words, the inputs of the $\emph{N}_{feat}$ are two images, and the output is the aggregated features $\bm{h_{agg}}$.

Many siamese CNN architectures can support the double inputs for similarity comparison \citep{8950457} or regression \citep{held2016learning}, whose weights are shared between the parallel streams. However, because the motion is specifically extracted defined in Eq.~(\ref{motion}), \textcolor{black}{$\bm{I_{df}}$ is different from the original frame $\bm{I_{f}}$.} Considering \textcolor{black}{that} they lie on different manifolds, the siamese CNN is not suitable for our task. We design a novel pseudo-siamese network \citep{hughes2018identifying} to handle this issue. 

First, the pseudo-siamese network is used to extract similar features between $\bm{I_{f}}$ and $\bm{I_{df}}$. The pseudo-siamese network can ensure the size and channel are completely equal at the same level layers. The architectures of the two sub-networks are totally the same, while the weights are learned separately. In Fig.~\ref{fig1}, the front network (FN) corresponds to the front difference network (FDN) while the back network (BN) corresponds to the back difference network (BDN). Then, $\emph{N}_{feat}=\{ \emph{N}_{FN},\emph{N}_{FDN},\emph{N}_{BN},\emph{N}_{BDN}\}$. The shape of features $\bm{h_{fn}}$ is equal to the features $\bm{h_{fdn}}$'s:
\begin{equation}\begin{aligned}
&\bm{h_{fn}}=\emph{N}_{FN}(\bm{I_{f}}) \\
&\bm{h_{fdn}}=\emph{N}_{FDN}(\bm{I_{df}}). \\
\end{aligned}
\end{equation}

Second, to improve the accuracy and robustness of human head detection, the network should focus on moving areas that contain human heads. To enhance the motion regions and suppress the non-moving regions, we present shallow features aggregation (SFA) and deep features aggregation (DFA). The goal of the SFA is to aggregate the shallow features at the low-level layers. To reserve more temporal information, MPSN directly learns the sum of \textcolor{black}{the} two features after activation functions:
\begin{equation}
SFA:\bm{h_{fn}}=\bm{h_{fdn}}+\bm{h_{fn}}.
\end{equation}

Intuitively, the motion regions are added to the original features through this simple low-level aggregation. SFA can enhance the small and moving pixels but may be inaccurate when similar background objects exist. In fact, \textcolor{black}{it is hard for SFA} to ensure the activation scores of human heads are higher than background objects'. To further enhance the moving instances and suppress the static background instances, we conduct the DFA on the position-sensitive score maps:
\begin{equation}\begin{aligned}
& \bm{h_{bn}}=\emph{N}_{BN}(\bm{h_{fn}}) \\
& \bm{h_{bdn}}=\emph{N}_{BDN}(\bm{h_{fdn}}) \\
& DFA:\bm{h_{agg}}=\alpha \cdot \bm{h_{bn}} \odot \bm{\sigma(h_{bdn})}+\beta \cdot \bm{h_{bdn}}. \\
\end{aligned}
\label{DFA}
\end{equation}

In the deep layers, the instance-level features can represent head activation scores. The feature value of $\bm{h_{bdn}}$ \textcolor{black}{represents} the spatial intensity of moving heads. At every channel, we resize $\bm{h_{bdn}}$ with the sigmoid functions to $(0,1)$. The scores are multiplied with the features $\bm{h_{bn}}$ by element-wise (Hadamard) product in Eq.~\eqref{DFA}. In other words, the $\bm{\sigma(h_{bdn})}$ can be seen as a special mask to activate the local regions of the original features maps $\bm{h_{bn}}$. Unlike normal masks in image processing, this mask is dense and learnable: every value belongs to $(0,1)$ and is learned from the above pseudo-siamese network. This simple operator is powerful because the moving instances will be partly reserved while the static instances will be suppressed.

Note that $\beta \cdot \bm{h_{bdn}}$ is added at the Eq.~\eqref{DFA}. Because the features $\bm{h_{bdn}}$ are resized to $(0,1)$ and the features $\bm{h_{bn}}$ are normalized (e.g., Batchnorm, Instancenorm, Layernorm) before activation functions, the activated head features will be decreased. To compensate for this loss and enhance the regions of moving heads, we add $\bm{h_{bdn}}$ after the element-wise product in Eq.~\eqref{DFA}.

\subsection{Detection network}
\textcolor{black}{
After obtaining the aggregated feature $\bm{h_{agg}}$, we use a detector to get head proposals.
\begin{equation}
\label{output}
(\bm{c,t})=\emph{N}_{det}(\bm{h_{agg}})
\end{equation}
where output $\bm{t}=(\bm{t_{x}},\bm{t_y},\bm{t_w},\bm{t_h})$ is a matrix that contains the parameterized coordinates of anchor boxes. $\bm{t_{x}},\bm{t_y}$ are the center coordinates of boxes, while $\bm{t_w},\bm{t_h}$ are the width and height of boxes. $\bm{c}$ is another matrix that contains class probabilities of anchor boxes. In the right of Fig.~\ref{fig1}, our method will predict 5700 anchor boxes (50x38x3) in each image. $\bm{c}$ is marked as the yellow block, while $\bm{t}$ is marked as the blue block. Because the scales of heads are diverse, we select three anchors of size 16x16, 32x32, and 64x64 to cover most heads. For each anchor, our method will predict 4 regression coordinates and 2 probability scores. Thus, the dimension of $\bm{t}$ is 5700x4, while the dimension of $\bm{c}$ is 5700x2.}

\textcolor{black}{On the other hand, to train the proposed network, the labels of head boxes need to be generated. We convert each ground-truth head box $G=(G_{x},G_{y},G_{w},G_{h})$ to parameterized coordinates $\hat{t}=(\hat{t_{x}},\hat{t_{y}},\hat{t_{w}},\hat{t_{h}})$, through the following equation:
\begin{equation} 
\begin{aligned}
\hat{t_{x}} &=\left(G_{x}-B_{x}\right) / B_{w} \\
\hat{t_{y}} &=\left(G_{y}-B_{y}\right) / B_{h} \\
\hat{t_{w}} &=\log \left(G_{w} / B_{w}\right) \\
\hat{t_{h}} &=\log \left(G_{h} / B_{h}\right),
\end{aligned}
\end{equation}
where $B=(B_{x},B_{y},B_{w},B_{h})$ is one of the base anchors generated throughout the image \citep{6909475}. By this equation, we can get the ground truth of coordinates $\bm{\hat{t}}$. On the other hand, the ground truth of classification $\bm{\hat{c}}$ can be generated directly by judging whether the anchor contains a head, i.e., 1 or 0.}

\textcolor{black}{Following the idea of RPN, we employ a multi-task loss function, as shown in Eq.~(\ref{loss}).
\begin{equation} \label{loss}
y_f=\emph{L}(\bm{c,t})=\frac{1}{N_s}\sum_{i}[\emph{L}_{cla}(\bm{c_i},\bm{\hat{c_i}})+\emph{L}_{box}(\bm{t_i},\bm{\hat{t_i}})].
\end{equation}
For each bounding box, $\hat{c}$ is the ground truth of classification while $\hat{t}$ is the ground truth of coordinates, as described above. $i$ is the index of the bounding box. We expect to minimize the distance between each predicted anchor $(\bm{c_i},\bm{t_i})$ and the corresponding ground truth $(\bm{\hat{c_i}},\bm{\hat{t_i}})$. $\emph{L}_{cla}$ is the cross-entropy loss while $\emph{L}_{box}$ is the smooth $L_{1}$ loss. The loss terms are normalized by $N_s$, which is the number of positive samples. Moreover, the loss terms are computed only using positive samples. The positive samples are defined by three strategies:(1) An anchor which has Intersection over Union (IoU) overlap \citep{6909475} with ground truth  $\geq$ 0.7 is a labeled positive sample. (2) Each ground truth has many IoUs with anchors, and we label the anchors corresponding to the maximum IoU as positive samples. (3) We restrict the number of positive samples $\leq$ 16.}

\textcolor{black}{After defining the loss function, the backbone network and detector will be jointly trained end-to-end. The training hyperparameters are presented in Appendix.}

\begin{table}[t]
\caption{Comparison of MPSN against other methods on Brainwash dataset and Restaurant dataset.}
\vspace{-1em}
\begin{center}
\begin{tabular*}{\tblwidth}{@{}LLLL@{}}
\toprule
Dataset&Method&Backbone&$AP_{50}$\\
\midrule 
\multirow{6}*{Brainwash}& ReInspect &GoogLeNet&$0.78$\\

~&SSD &ResNet101&$0.80$\\

~&DETR &ResNet18&$0.53$\\

~&FCHD&VGGNet16&$0.70$\\
~&\textcolor{black}{CrowdDet}&\textcolor{black}{ResNet50}&\textcolor{black}{$0.67$}\\
~&\textcolor{black}{Iter-E2EDET}&\textcolor{black}{ResNet50}&\textcolor{black}{$0.52$}\\
~&MPSN& MobileNetv2 &$\bm{0.90}$\\
~&MPSN& VGGNet16 &$\bm{0.91}$\\
\hline
\multirow{7}*{Restaurant}& YOLOX &CSPNet&$0.83$\\
~&HTC++ &Swin-B&$0.68$\\
~&SSD &ResNet18& $0.51$\\
~&FCHD&VGGNet16&$0.75$\\
~&\textcolor{black}{CrowdDet}&\textcolor{black}{ResNet50}&\textcolor{black}{$0.61$}\\
~&\textcolor{black}{Iter-E2EDET}&\textcolor{black}{ResNet50}&\textcolor{black}{$0.61$}\\
~&MPSN& MobileNetv2 &$\bm{0.84}$\\
~&MPSN&VGGNet16&$\bm{0.86}$\\
\bottomrule
\end{tabular*}
\vspace{-1.7em}
\label{table2}
\end{center}
\end{table}


\section{\textcolor{black}{EXPERIMENTS and DISCUSSION}}

First, we compare MPSN with various methods on two public indoor crowd datasets in Section \ref{Datasets}. Considering \textcolor{black}{that} SFA, DFA can be embedded in most CNN-based models, we mainly focus on the flexibility and comparison study (in Section \ref{Comparison with baselines}, \ref{Flexibility of MPSN}, and \ref{Diffabs versus Flow}). Based on different backbone networks and detectors, we show that MPSN can recover missed heads and filter the false background positives, giving a significant performance boosting. After comparison, we choose the best model to count indoor occupants, balancing the low computational complexity and high accuracy (in Section \ref{OCC}). In addition, adversarial samples are implemented to evaluate the robustness of MPSN (in Section \ref{Robustness to adversarial samples}). Finally, we also provide the mathematical solution of small perturbations to select more robust models.

\subsection{Datasets}\label{Datasets}

To evaluate MPSN, we test the publicly available Brainwash crowd dataset \citep{DBLP:journals/corr/StewartA15} and Restaurant dataset \citep{9150687}. For evaluation metrics, we use the standard average precision ($AP_{50}$). The Brainwash dataset includes 11917 images with 91146 labeled people, in which the test set contains 484 images. The Restaurant dataset was collected in four different indoor locations at a restaurant. It includes 1610 images, from which the test set contains 123 images. The images are extracted from the video with a large time interval, thus having significant diversity and difference.
Fortunately, the camera \textcolor{black}{was} stable, and therefore the proposed motion estimation method is applicable.

\subsection{Comparison with baselines}\label{Comparison with baselines}

\textcolor{black}{To compare to the state-of-the-art (SOTA) methods}, we \textcolor{black}{tested the AP of} different methods on two datasets in Table~\ref{table2}. The detailed network architectures and hyperparameters are \textcolor{black}{presented} in Appendix.

The authors who published the Brainwash dataset \citep{DBLP:journals/corr/StewartA15} provided a baseline detection method named ReInspect. SSD \citep{Zhang_2018_CVPR} algorithm uses multi-scale feature maps to achieve high detection accuracy. FCHD is a fast and accurate head detector with a specific anchor size selection strategy. HTC++ framework \citep{DBLP:journals/corr/abs-1901-07518} combines detection and segmentation tasks into a joint multi-stage processing and utilizes spatial context to further boost the performance.
YOLOX \citep{DBLP:journals/corr/abs-2107-08430} is a high-performance anchor-free detector through integrating YOLO series. \textcolor{black}{CrowdDet \citep{Chu_2020_CVPR} is a SOTA detector that achieves 0.907 AP performance in the challenging CrowdHuman dataset.} To simplify detection pipelines and bypass surrogate tasks, DETR directly utilizes the transformer for object sequence prediction. \textcolor{black}{Iter-E2EDET \citep{2022arXiv220307669Z} is a SOTA people detector and achieves the highest accuracy in the CrowdHuman detection task \footnote{https://paperswithcode.com/sota/object-detection-on-crowdhuman-full-body} so far. }

Compared with the prior methods, our proposed method, MPSN, \textcolor{black}{achieved the best performance accuracy in both datasets}. Considering \textcolor{black}{that} the MobileNetv2 backbone \citep{DBLP:journals/corr/abs-1801-04381} has fewer parameters and lower computational complexity, we \textcolor{black}{used} it as the default backbone in further discussion.

\subsection{Flexibility of MPSN}\label{Flexibility of MPSN}

To sufficiently evaluate the flexibility, we \textcolor{black}{conducted} three experiments based on FCHD. Table~\ref{flexible} summarizes the performance on two datasets. We \textcolor{black}{selected} the best model on the evaluation dataset ($val AP_{50}$) and \textcolor{black}{tested} this model on the test dataset ($test AP_{50}$). 

\begin{table}[htp]
\caption{Comparison of MPSN against the single frame and two frames method among different backbones on two datasets.}
\vspace{-1em}
\begin{center}
\begin{tabular*}{\tblwidth}{@{}LLLLL@{}}
\toprule
Dataset&Method&VGGNet16&MobileNetv2&ResNet18\\
\midrule 
\multirow{3}*{Brainwash}& Single frame &$0.876$&$0.828$&$0.855$\\

~&Two frames &$0.868$&$0.821$&$0.857$\\

~& MPSN &$\bm{0.878}$&$\bm{0.890}$&$\bm{0.885}$\\
\hline
\multirow{3}*{Restaurant}& Single frame &$0.750$&$0.785$&$0.769$\\
~&Two frames &$0.782$&$0.808$&$0.768$\\
~& MPSN &$\bm{0.857}$&$\bm{0.838}$&$\bm{0.802}$\\
\hline
\multicolumn{2}{l}{Parameters}&$123.8$M&$\bm{6.7}$M&$24.9$M\\
\multicolumn{2}{l}{Inference time}&$149$ms&$125$ms&$\bm{109}$ms\\
\bottomrule
\end{tabular*}
\vspace{-1.7em}
\label{flexible}
\end{center}
\end{table}

\begin{itemize}
  \item Single frame: training the general model with single frame as the input.
  \item Two frames: training the pseudo-siamese model with two adjacent frames ($\bm{I_{f}},\bm{I_{f-1}}$). In this situation, the DFA degenerates to $\bm{h_{agg}}=\bm{h_{bn}}+\bm{h_{bdn}}$.
  \item Diffabs: using pre-computed sequence pairs ($\bm{I_{f}},\bm{I_{df}}$) to train the pseudo-siamese model with the $f$-th frame ground truth.
  \item DFA: two variants of DFA are considered, where $\bigodot$ represents using our Eq.~\eqref{DFA} to aggregate the deep features, and $\bigoplus$ represents using the simple addition.
\end{itemize}

We \textcolor{black}{tested MPSN with} different backbone networks (VGGNet16, MobileNetv2, Resnet18). In each backbone network and dataset, we \textcolor{black}{fixed} hyperparameters to fairly compare single frame, two frames, and MPSN methods. The results in Table~\ref{flexible} show MPSN \textcolor{black}{achieved} superior performance on two datasets. It means that MPSN does not rely on fixed network architectures, thus it can be flexibly applied in each CNN-based detection framework with high accuracy. \textcolor{black}{We also compare computational complexity of MPSN with different backbone networks in Table~\ref{flexible}. It can be seen that, MPSN with MobileNetv2 has smaller parameters while MPSN with ResNet18 has faster inference speed.}

As for "MobileNetv2" on the Restaurant dataset, the single frame method \textcolor{black}{achieved} a reasonable performance of 0.785 AP. Simultaneously, the two frames method \textcolor{black}{achieved} similar accuracy of 0.808 AP. After applying MPSN, we achieve \textcolor{black}{an AP of} 0.838 with the same hyperparameters. This significant improvement means that MPSN can guide CNN to extract motion features effectively, resulting in higher accuracy.

To verify the flexibility, we also \textcolor{black}{applied our MPSN to} DETR framework, which detects objects using a transformer. The DETR with ResNet18 is compared in Table~\ref{DETR}. In our experiments, we only \textcolor{black}{added} DFA in the ResNet18 backbone network before the transformer decoder. The Diffabs and two frames methods (0.713 AP) \textcolor{black}{were} significantly improved. It means that the additional information can help the performance. Importantly, compared with the two frames situation at the same scale of model parameters, our Diffabs and DFA method \textcolor{black}{achieved} 0.762 AP, about 6.87$\%$ improvement. It benefits from the crafty design of DFA and the effective use of motion information. However, the transformer decoder has more parameters, which \textcolor{black}{makes it} not applicable in practice. In contrast, since the MobileNetv2 backbone network has fewer parameters and computational complexity \citep{DBLP:journals/corr/abs-1801-04381}, it is suitable for edge deployments.

\begin{table}[t]
\caption{Ablation with DETR on the Brainwash dataset.}
\vspace{-1em}
\begin{center}

\begin{tabular*}{\tblwidth}{@{}LLLLLL@{}}
\toprule
Input&SFA&DFA&$val AP_{50}$&$test AP_{50}$&Params\\
\midrule

Single frame &&& $0.509$ &$0.525$&$344.9$M\\

Two frames &&$\bigoplus$&$0.783$ &$0.713$&$477.8$M\\

Diffabs&&$\bigoplus$&$0.723$ &$0.713$&$389.7$M\\

Diffabs&&$\bigodot$&$0.774$ &$\bm{0.762}$&$477.8$M\\
\bottomrule
\end{tabular*}
\vspace{-1.7em}
\label{DETR}
\end{center}
\end{table}

\begin{figure*}[htbp]
 \begin{minipage}{0.19\linewidth}
  \centerline{\includegraphics[width=\textwidth]{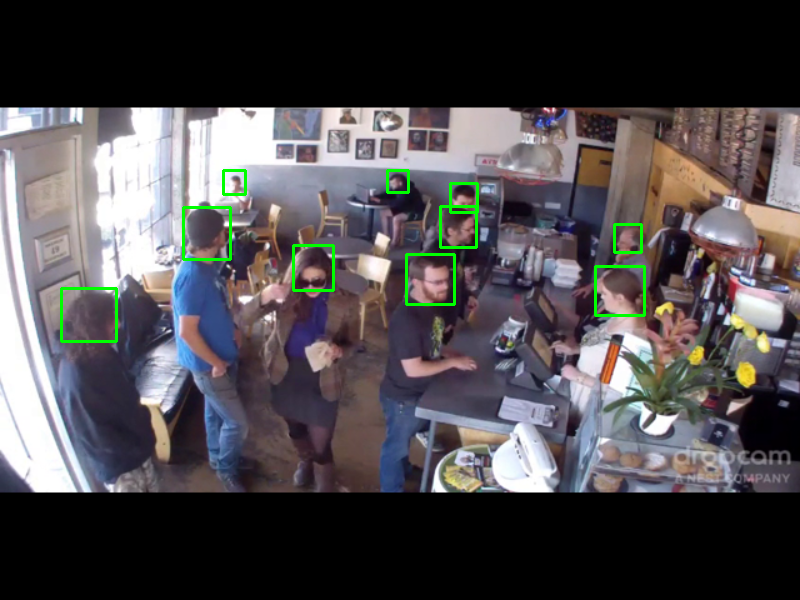}}
  \vspace{3pt}
  \centerline{\includegraphics[width=\textwidth]{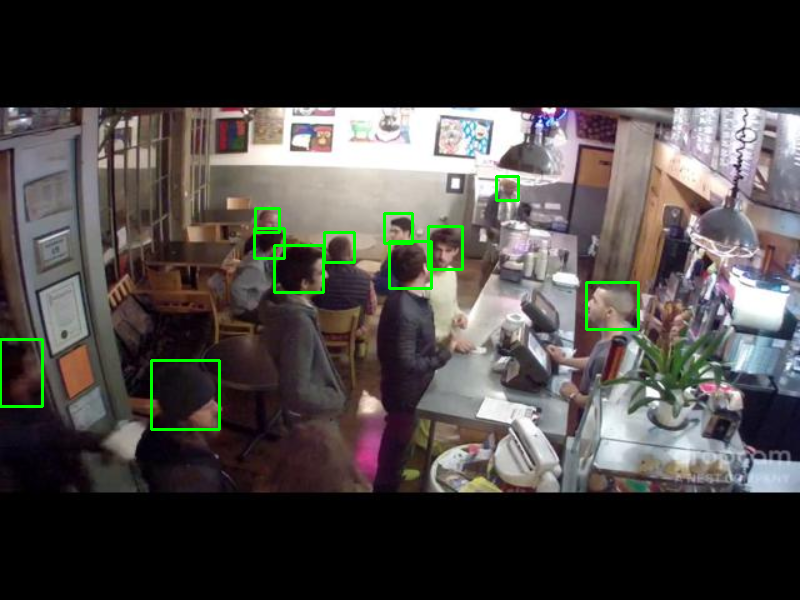}}
 \end{minipage}
 \begin{minipage}{0.19\linewidth}
  \centerline{\includegraphics[width=\textwidth]{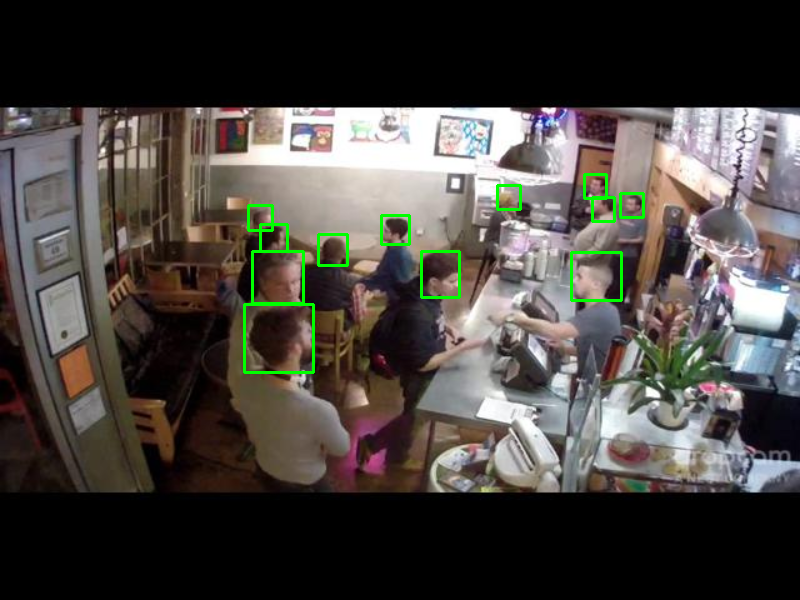}}
  \vspace{3pt}
  \centerline{\includegraphics[width=\textwidth]{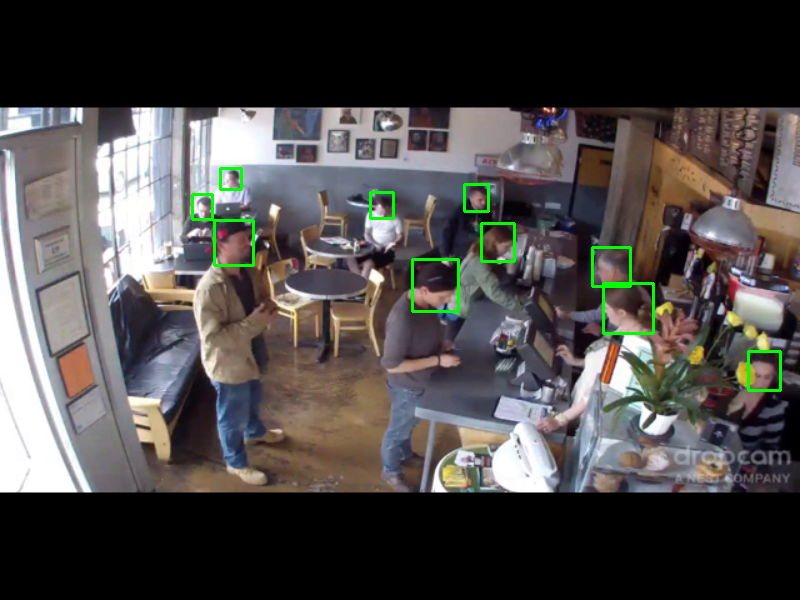}}
 \end{minipage}
 \begin{minipage}{0.19\linewidth}
  \centerline{\includegraphics[width=\textwidth]{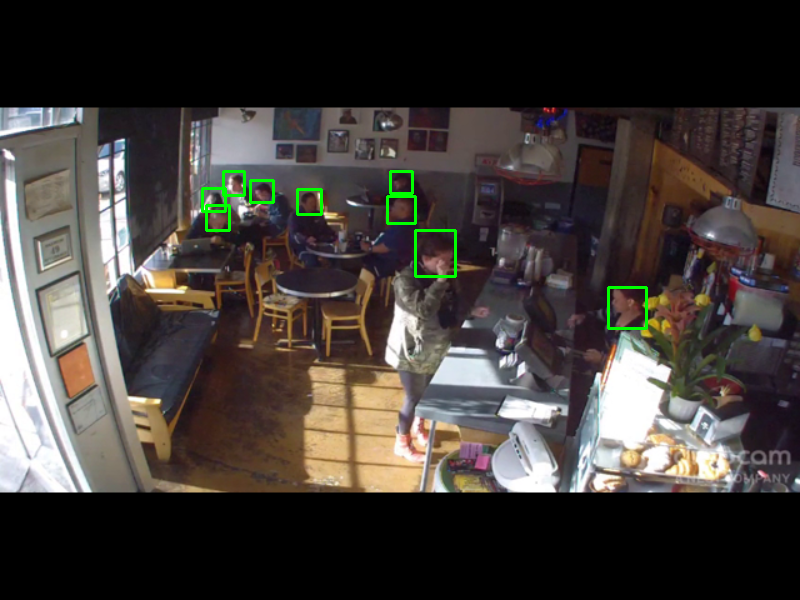}}
  \vspace{3pt}
  \centerline{\includegraphics[width=\textwidth]{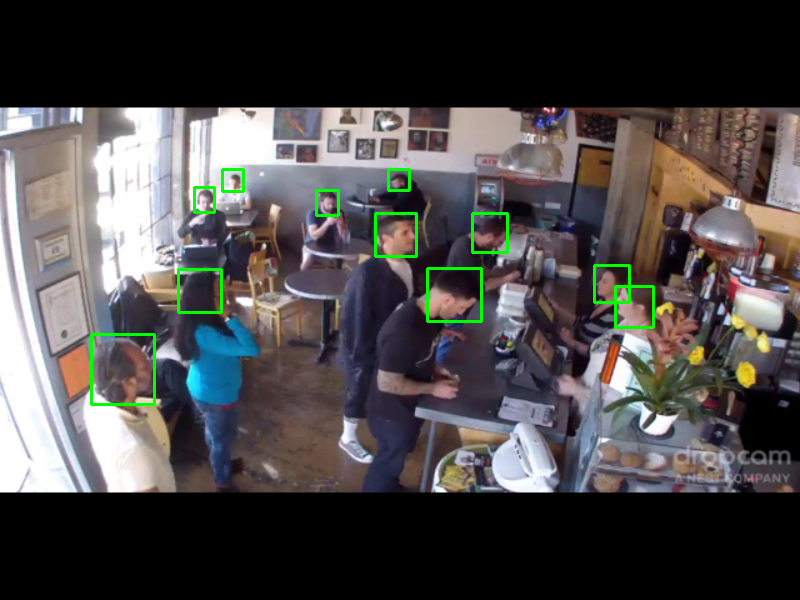}}
 \end{minipage}
 \begin{minipage}{0.19\linewidth}
  \centerline{\includegraphics[width=\textwidth]{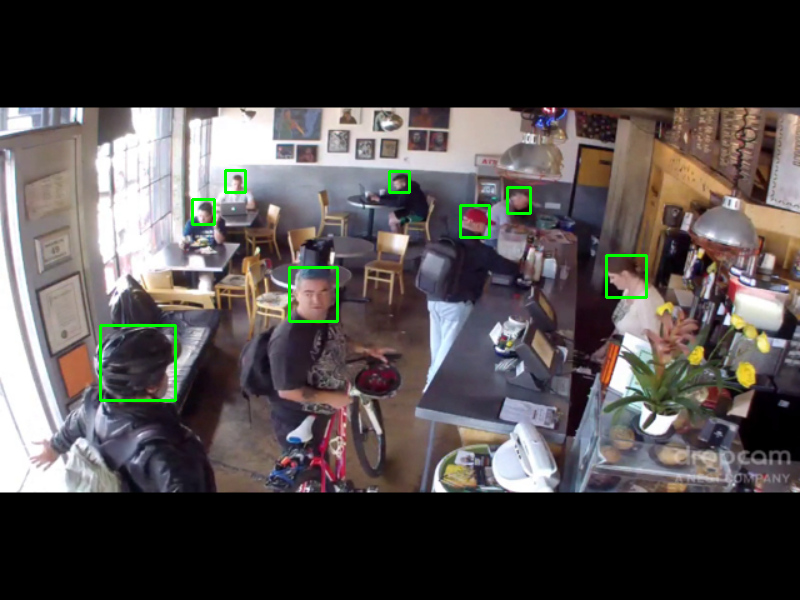}}
  \vspace{3pt}
  \centerline{\includegraphics[width=\textwidth]{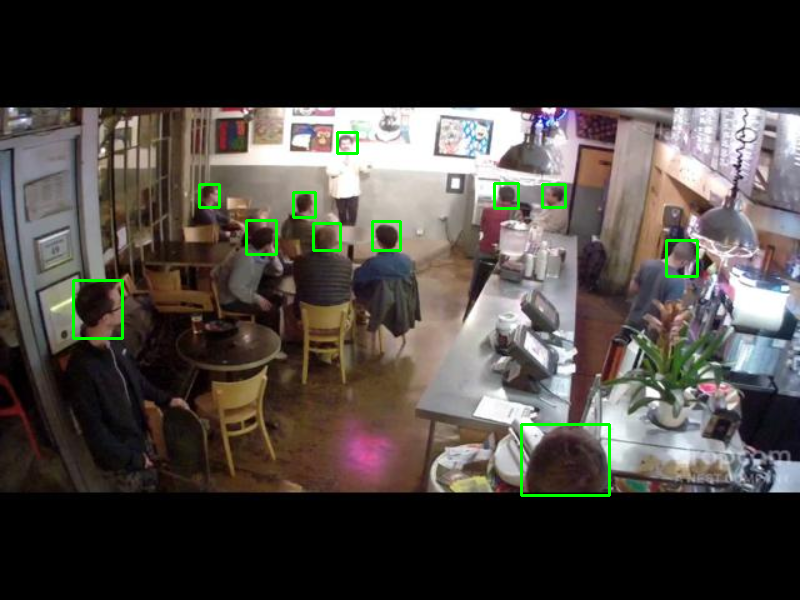}}
 \end{minipage}
 \begin{minipage}{0.19\linewidth}
  \centerline{\includegraphics[width=\textwidth]{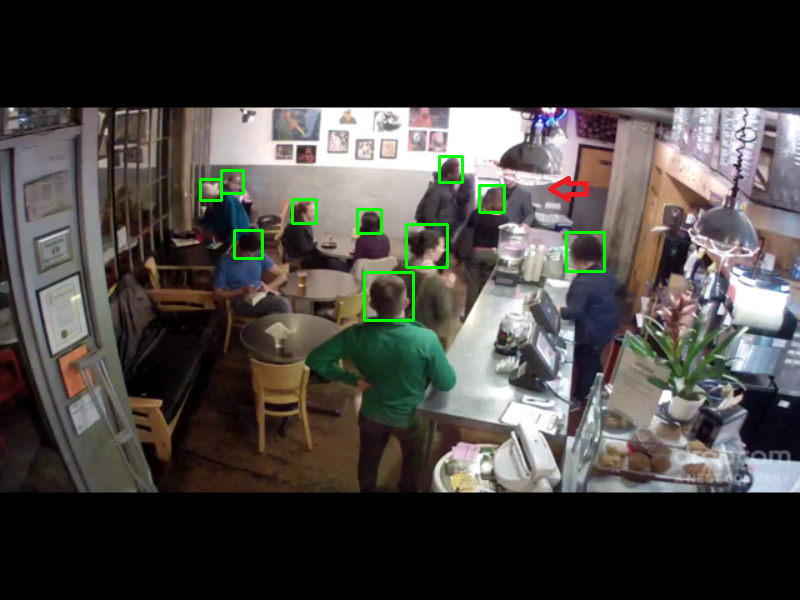}}
  \vspace{3pt}
  \centerline{\includegraphics[width=\textwidth]{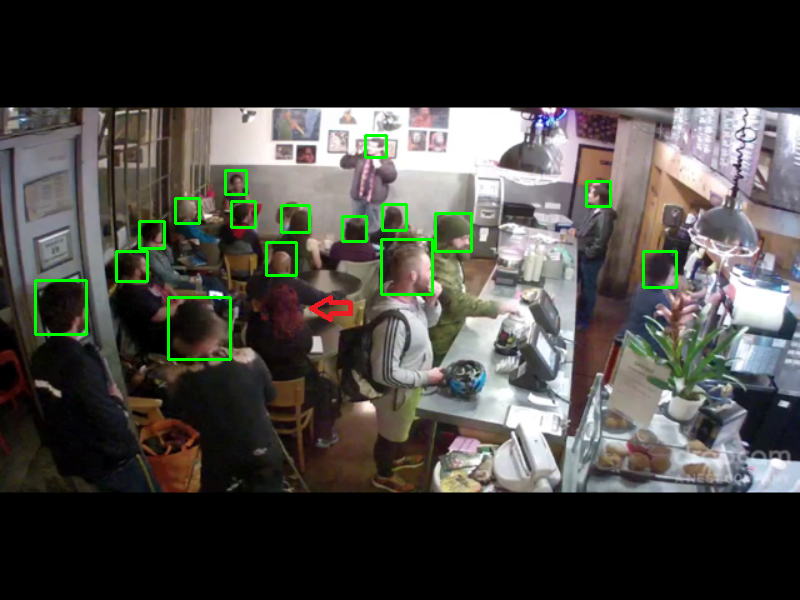}}
 \end{minipage}
 \begin{minipage}{0.19\linewidth}
  \centerline{\includegraphics[width=\textwidth]{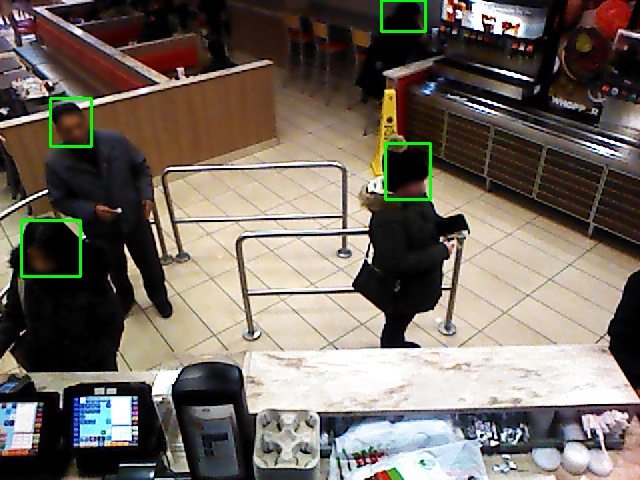}}
  \vspace{3pt}
  \centerline{\includegraphics[width=\textwidth]{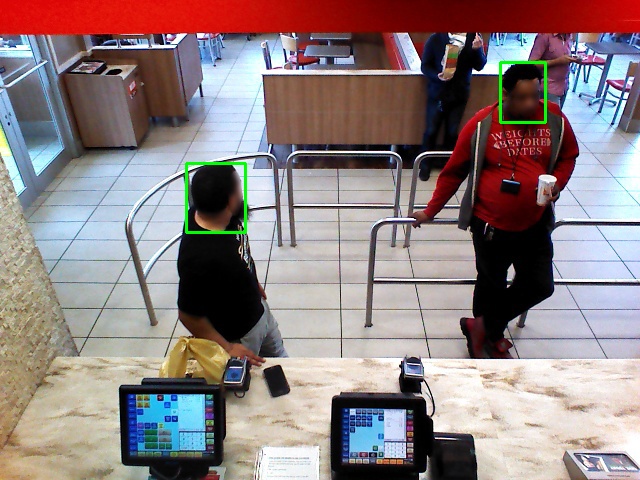}}
 \end{minipage}
 \begin{minipage}{0.19\linewidth}
  \centerline{\includegraphics[width=\textwidth]{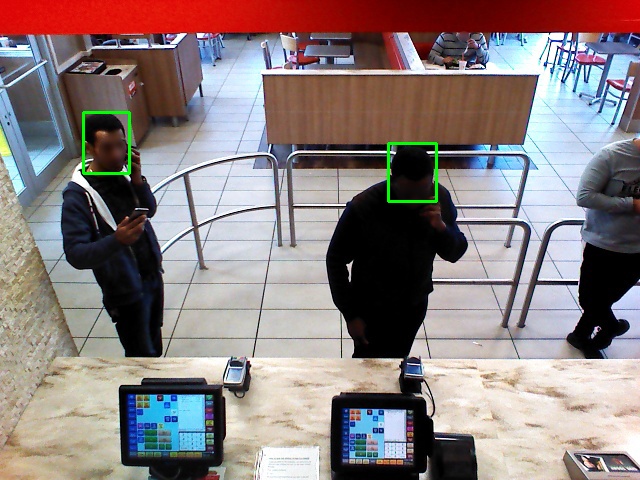}}
  \vspace{3pt}
  \centerline{\includegraphics[width=\textwidth]{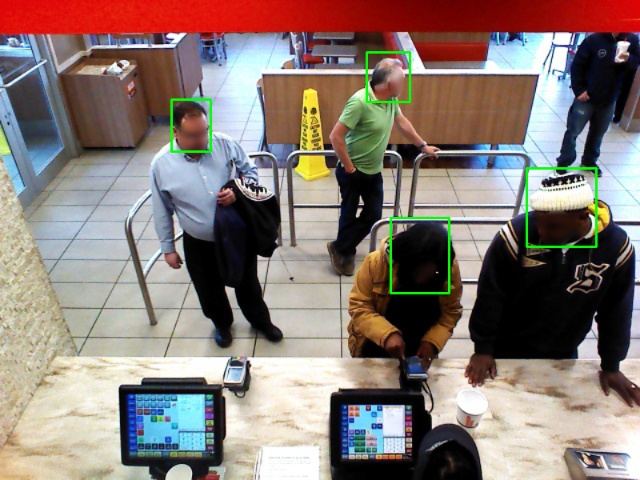}}
 \end{minipage}
 \begin{minipage}{0.19\linewidth}
  \centerline{\includegraphics[width=\textwidth]{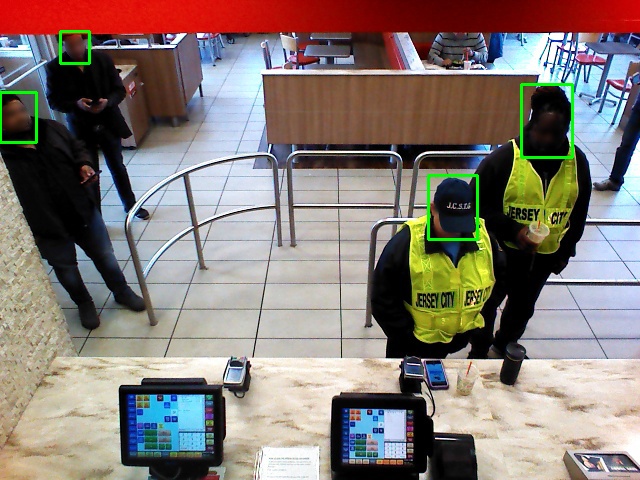}}
  \vspace{3pt}
  \centerline{\includegraphics[width=\textwidth]{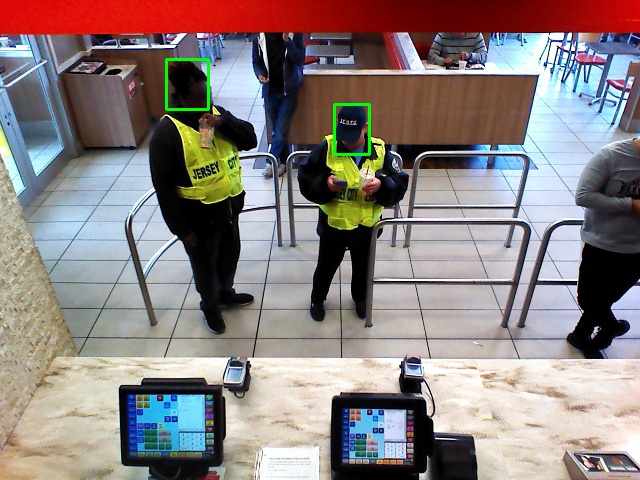}}
 \end{minipage}
 \begin{minipage}{0.19\linewidth}
  \centerline{\includegraphics[width=\textwidth]{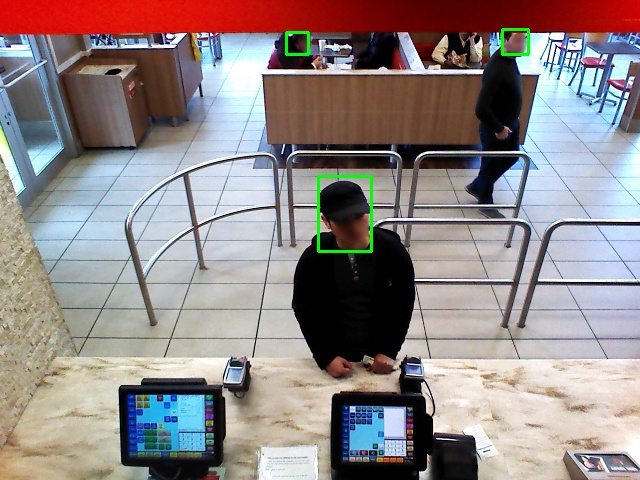}}
  \vspace{3pt}
  \centerline{\includegraphics[width=\textwidth]{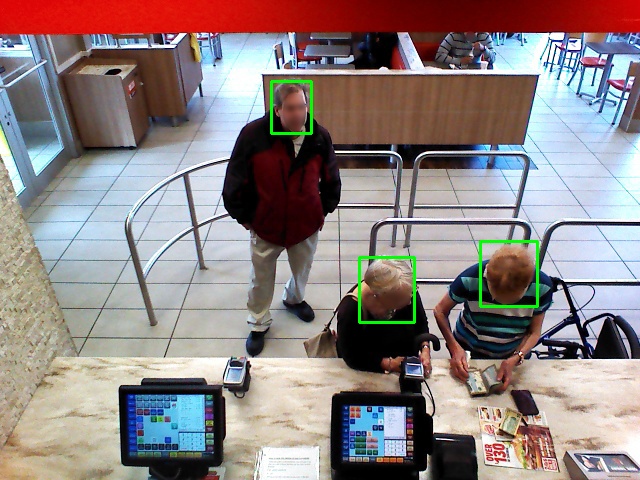}}
 \end{minipage}
 \begin{minipage}{0.19\linewidth}
  \centerline{\includegraphics[width=\textwidth]{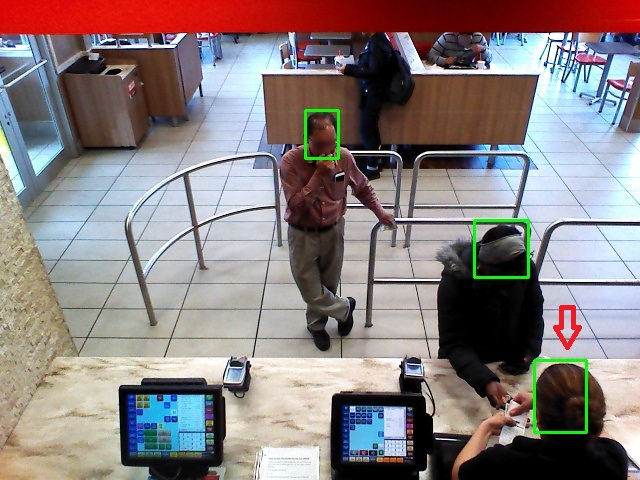}}
  \vspace{3pt}
  \centerline{\includegraphics[width=\textwidth]{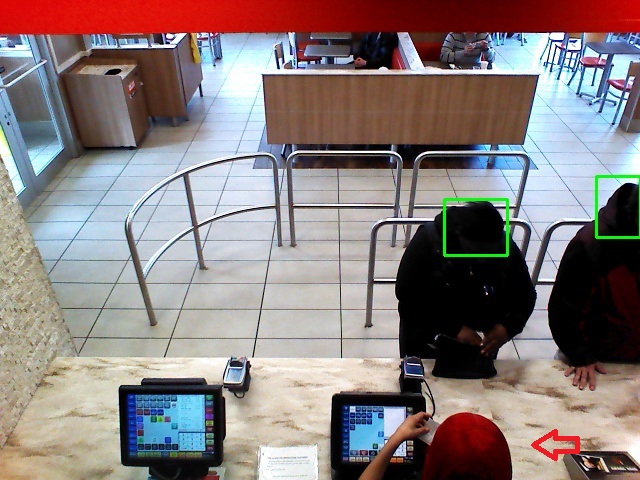}}
 \end{minipage}
\caption{Qualitative results of MPSN on test set of Brainwash dataset and Restaurant dataset. In the first four columns, we present successful results where all human heads are detected accurately even though several heads are tiny. In the last column, we provide some failure instances. \textcolor{black}{As marked by red arrows in the pictures, four heads are missed. In the top-right figure, the man is occluded by a bright lamp and another person. In the second right figure, the back-view woman with red hair is missed because her hair color is close to near pixels. They are hard to be discovered even with human eyes. In the third right figure, the head is not located accurately. In the bottom right figure, the person is missed. Two potential reasons: only part head can be seen; heads with red cap in datasets are rare.}}
\label{Qualitative}
\end{figure*}

\subsection{Diffabs versus Flow}\label{Diffabs versus Flow}

\begin{table}[h]
\caption{Comparison of Diffabs against Flow on the two datasets.}
\vspace{-1em}
\begin{center}

\begin{tabular*}{\tblwidth}{@{}LLLLLLL@{}}
\toprule

\multirow{2}{*}{Dataset}&\multirow{2}{*}{Network}&\multicolumn{2}{c}{Flow}&\multicolumn{2}{c}{Diffabs}\\
\cline{3-6}

&&$val AP_{50}$&$test AP_{50}$&$val AP_{50}$&$test AP_{50}$\\
\hline

\multirow{3}*{Restaurant}&VGGNet16& $0.789$&$0.812$&$\bm{0.816}$&$\bm{0.857}$\\
~&MobileNetv2& $\bm{0.759}$&$0.790$&$0.752$&$\bm{0.838}$\\
~&Resnet18& $0.743$&$0.793$&$\bm{0.747}$&$\bm{0.802}$\\
\hline
\multirow{3}*{Brainwash}&VGGNet16& $0.866$&$0.903$&$\bm{0.884}$&$\bm{0.909}$\\
~&MobileNetv2& $0.865$&$0.880$&$\bm{0.867}$&$\bm{0.899}$\\
~&Resnet18& $0.850$&$0.880$&$\bm{0.852}$&$\bm{0.885}$\\
\bottomrule
\end{tabular*}
\vspace{-1.7em}
\label{Diffabs}
\end{center}
\end{table}

Optical flow is widely used in video analysis and processing. We compare optical flow and frame difference in this part. We \textcolor{black}{used} SOTA optical flow estimation method named RAFT \citep{teed2020raft} to extract flow images, which are the inputs to our MPSN. The frame difference \textcolor{black}{was} simply implemented by subtraction and absolute value operators. In this experiment, we \textcolor{black}{fixed} DFA, and hyperparameters to compare Diffabs and Flow fairly. The results are shown in Table~\ref{Diffabs}. Diffabs \textcolor{black}{achieved} better performance than Flow in most situations, regardless of datasets and backbone networks. Note that the SOTA optical flow estimation method has more complex computation and more parameters than the simple frame difference method. In conclusion, as compared to Flow-based methods, the Diffabs-based methods have lower computational requirements while enjoying higher accuracy.

\subsection{An application: occupancy counting} \label{OCC}
An application of MPSN is occupancy counting, which is the core component of occupancy-based control in buildings. As a critical perception algorithm, MPSN provides the number of occupants, which can be applied to control the HVAC and lighting systems. For example, energy can be effectively saved and indoor air environment quality can be improved, through adjusting the opening rate of air dampers dynamically according to the number of people \textcolor{black}{in the building}. In this experiment, we \textcolor{black}{applied} MPSN to evaluate the performance of indoor occupancy counting. (In practice, balancing the low computational complexity and high accuracy, we \textcolor{black}{chose} the MobileNetv2 backbone as the test model.)

We define two evaluation indicators for occupancy counting as follows:

\begin{equation}\label{nmae}
\begin{aligned}
&\bm{NMAE}=\frac{1}{N}\sum_{\bm{f}=1}^N \frac{|n_f-\overline{n_f}|}{n_f+\overline{n_f}}, \\
&\bm{SCORE}=\frac{1}{N}\sum_{\bm{f}=1}^N (1-{\rm sign}|n_f-\overline{n_f}|),
\end{aligned}
\end{equation}
where $n_f$ represents the true number of people (ground truth) in the $f_{th}$ frame. And $\overline{n_f}$ represents the predicted number of people (predicted value) in the $f_{th}$ frame. The ground truth of the video is obtained manually, while the predicted value is output by MPSN. $sign$ represents the sign function. 

The \textcolor{black}{normalized mean absolute error} ($NMAE \in [0,1]$) is calculated through the absolute difference between $n_f$ and $\overline{n_f}$. We normalized it because the occupancy counting error increases with the number of people \citep{SUN2020109965}. SCORE (hit rate) is the ratio of ground truth equal to the predicted value. We expect NMAE to be smaller and SCORE to be bigger.
\textcolor{black}{Note that the occupancy counting application mainly focuses on the number of occupants in buildings. Thus, we assume the detected occupancy boxes match the ground truth position in Eq.~\eqref{nmae}.}

\begin{table}[t]
\caption{Performance of occupancy counting.}
\vspace{-1em}
\begin{center}
\begin{tabular*}{\tblwidth}{@{}LLLLL@{}}
\toprule
Dataset&NMAE&SCORE& Avg head counting\\
\midrule
Restaurant &0.125&0.525 &2.549\\
Brainwash &0.125&0.186 &10.345\\
\bottomrule
\end{tabular*}
\vspace{-1.7em}
\label{counting}
\end{center}
\end{table}

In Table \ref{counting}, the Avg head counting is the average occupancy counting in the test dataset. NMAE does not change even though the average numbers of people are significantly different \textcolor{black}{in} the two datasets. It partially shows that our NMAE is insensitive to the scale of occupants. On the other hand, the small NMAE indicates \textcolor{black}{that} MPSN has a good performance in occupancy counting, as shown in Fig. \ref{Qualitative}. Intuitively, if the true number of people ($n_f$) is 5, with the 0.125 NMAE, the predicted number ($\overline{n_f}$) will be 4 or 6 statistically. This small error can be tolerated for coarse-grained building control. Note that MPSN is only \textcolor{black}{a} head detector, thus post-processing and pre-processing methods are able to cascaded to improve the counting performance. These processing methods (such as max counting algorithm \citep{choi2021application}, clustering analyzer \citep{RN475} and occupancy frequency histogram \citep{SUN2022111593}) are beyond the scope of this paper.

\subsection{Robustness to adversarial samples}\label{Robustness to adversarial samples}

\begin{figure}[t]
\centering
\includegraphics[scale=0.19]{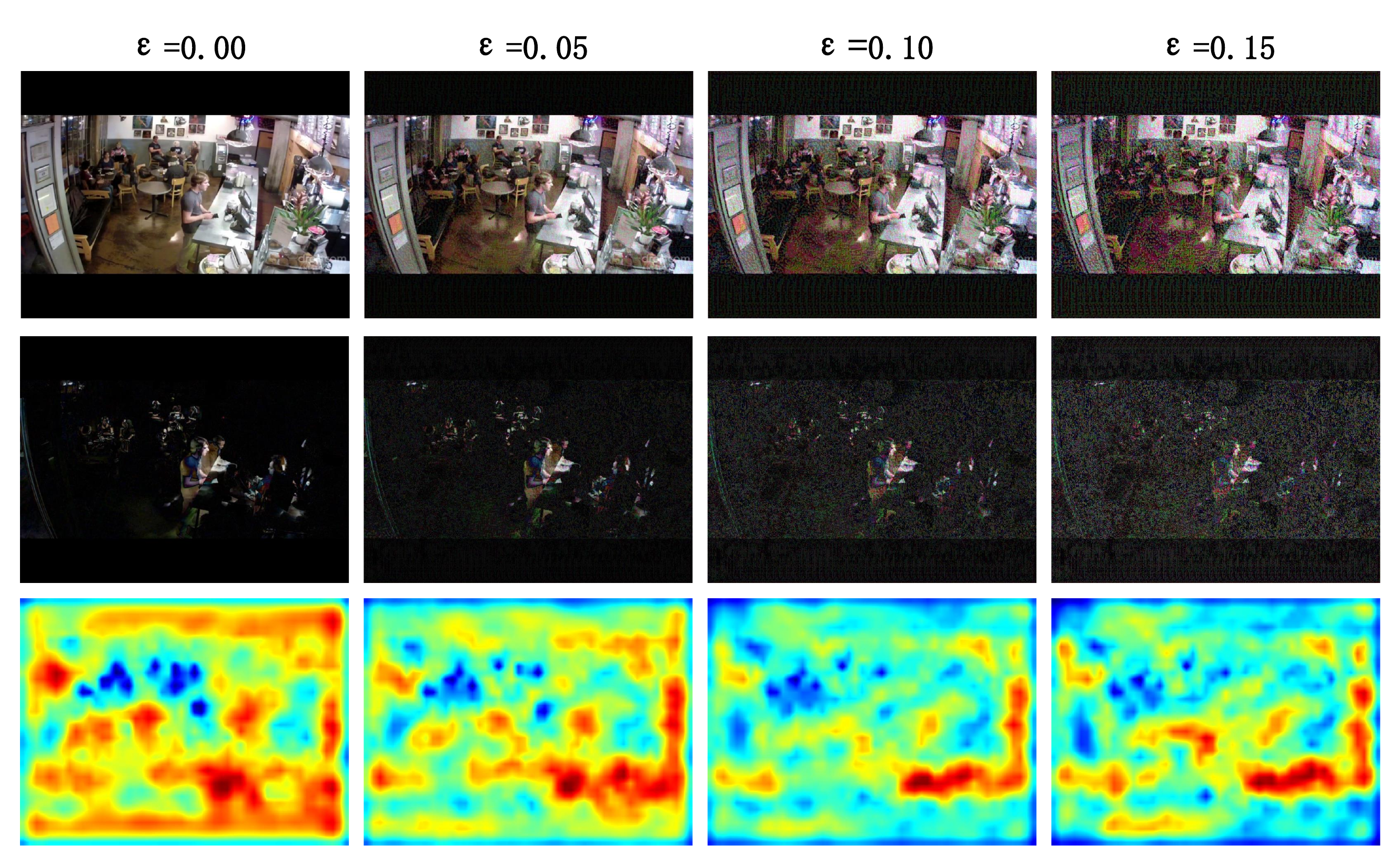}
\vspace{-2em}
\caption{Visualization of effects of increasing adversarial perturbation. First row represents adversarial images; second row represents motion in adversarial images; last row represents CAM heatmaps from the last CNN layer.}
\label{fig4}
\vspace{-1em}
\end{figure}

\begin{figure}[t]
\centering
\vspace{-1em}
\includegraphics[scale=0.5]{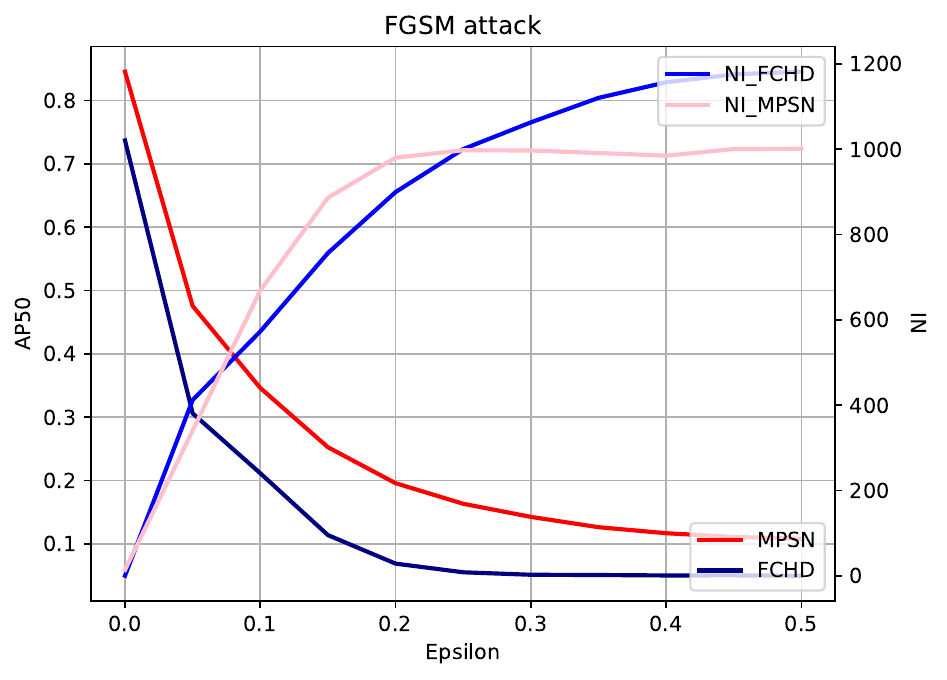}
\vspace{-0.5em}
\caption{Comparison of robustness between MPSN and FCHD under adversarial noise.}
\label{fig3}
\vspace{-1em}
\end{figure}

Recent studies \textcolor{black}{on} adversarial samples~\citep{akhtar2018threat} \textcolor{black}{found that} imperceptible perturbations in input images can completely fool the deep learning models. We \textcolor{black}{conducted} an experiment on robustness with adversarial samples. Instead of defending the adversarial attacks, the purpose of this experiment \textcolor{black}{was} to evaluate the robustness of MPSN. We \textcolor{black}{added} adversarial perturbations to our dataset, which are computed by the fast gradient sign method (FGSM) \citep{goodfellow2015explaining}:

\begin{equation}\label{attack input}
\begin{aligned}
&\bm{I_f^a}=\bm{I_f}+\epsilon {\rm sign}(\nabla_{\bm{I_f}} L(\bm{\theta},\bm{I_f},y_f)), \\
&\bm{I_{f-1}^a}=\bm{I_{f-1}}+\epsilon {\rm sign}(\nabla_{\bm{I_f-1}} L(\bm{\theta},\bm{I_{f-1}},y_{f-1})),
\end{aligned}
\end{equation}
where $\epsilon \in [0,1]$. In Fig.~\ref{fig3}, with the $\epsilon$ increasing, the attack effect also increases; thus the $AP_{50}$ decreases. It is worth noting that the $AP_{50}$ of MPSN is always higher than FCHD \textcolor{black}{for} different $\epsilon$. It means that no matter how great this attack is, MPSN can achieve a better performance compared with FCHD. At $\epsilon=0$ level, the attack effects disappear. At $\epsilon=0.4$ level, MPSN converges to higher accuracy. Detailed effects against adversarial perturbations are visualized in Fig.~\ref{fig4}. To some extent, the blue areas in heatmaps can represent the head positions. With the increasing perturbations, the heatmaps become inaccurate, which confirms the decrease of AP in Fig.~\ref{fig3}. However, it looks like that the second row (frame difference) is less affected by the perturbations because the motion area is significant. Perhaps it reduces the perturbations and results in higher accuracy in Fig.~\ref{fig3}.

We \textcolor{black}{used} the Non-I.I.D. Index (NI)~\citep{HE2021107383} to evaluate the robustness in Fig.~\ref{fig3}:
\begin{equation}\label{NI}
\begin{aligned}
&\bm{{NI_{MPSN}}}=\left\|\frac{\overline{g_\theta(\bm{I^a,I_{d}^a})}-\overline{g_\theta(\bm{I,I_{d}})}}{{\rm std}(g_\theta(\bm{I^a,I_{d}^a})\cup g_\theta(\bm{I,I_{d}}))}\right\|_2, \\
&\bm{{NI_{FCHD}}}=\left\|\frac{\overline{g_\varphi(\bm{I^a})}-\overline{g_\varphi(\bm{I})}}{{\rm std}(g_\varphi(\bm{I^a})\cup g_\varphi(\bm{I}))}\right\|_2,
\end{aligned}
\end{equation}
and
\begin{equation}
\begin{aligned}
\overline{g_\theta(\bm{I^a,I_{d}^a})} &=\frac{1}{N} \sum_{\bm{f}=1}^{N} g_\theta(\bm{I_{f}^a,I_{df}^a}), \\
\overline{g_\varphi(\bm{I^a})} &=\frac{1}{N+1}\sum_{\bm{f}=0}^N g_\varphi(\bm{I_f^a}),
\end{aligned}
\end{equation}
where $g_\theta$ represents the feature extractor of MPSN and $g_\varphi$ represents the feature extractor of FCHD. The superscript $\bm{a}$ represents the input image after attack, $\overline{(\cdot)}$ represents the first order moment, $\rm{std}$ is the standard deviation function and $\|\cdot\|_2$ represents the 2-norm. Eq.~\eqref{NI} measures the feature difference between original images and adversarial samples.

Since MPSN and FCHD have different network architectures, we slightly modified the NI as follows:

\begin{equation}\label{Ni final}
\begin{aligned}
&\bm{{NI_{MPSN}}}=\left\|\frac{\overline{CAM(\bm{h_{det}^{a}})}-\overline{CAM(\bm{h_{det}})}}{{\rm std}(CAM(\bm{h_{det}^{a}})\cup CAM(\bm{h_{det}}))}\right\|_2, \\
&\bm{{NI_{FCHD}}}=\left\|\frac{\overline{CAM(\bm{d_{det}^{a}})}-\overline{CAM(\bm{d_{det}})}}{{\rm std}(CAM(\bm{d_{det}^{a}})\cup CAM(\bm{d_{det}}))}\right\|_2,
\end{aligned}
\end{equation}
where $\bm{h_{det}}$ and $\bm{d_{det}}$ represent the features from the last CNN layer of MPSN and FCHD, respectively. The main advantage of $CAM$ is interpretability. Thus we specify the $CAM$ heatmap from the last CNN layer as $g(.)$ function.

The comparison results are shown in Fig.~\ref{fig3}. The strong correlation between NI and testing error \textcolor{black}{was} proved in \citep{HE2021107383}. It seems that FCHD is more robust than MPSN if $\epsilon \in [0.07, 0.25]$. It can be explained that the AP decrease of MPSN is serious. However, in applications, we focus on the small $\epsilon$ because the great perturbations violate small input changes \textcolor{black}{and} even destroy the original image \citep{feinman2017detecting}. According to Fig.~\ref{fig3}, if $\epsilon \leq 0.07$, MPSN is more robust and accurate than FCHD. In this situation, we are more likely to apply MPSN in real systems. On the other hand, if $\epsilon \geq 0.07$, we may choose FCHD. \textcolor{black}{Therefore, the threshold is very important in selecting} the more robust model and further guide the applications. While the threshold is experimental, we further discuss the mathematical solution of the small $\epsilon$ in the Appendix.

\section{CONCLUSIONS}
In this \textcolor{black}{study}, we present MPSN to suppress false positives and recover missed detection in video head detection. We design SFA and DFA by introducing motion information, guiding the pseudo-siamese model to extract robust head motion features. Experimental results show that MPSN achieves higher performance on two challenging datasets. Experiments based on different backbone networks (VGGNet16, MobileNetv2, Resnet18) verify the flexibility of MPSN. Compared with optical flow-based methods, our Diffabs-based methods have lower computational requirements and higher accuracy in experiments. Adversarial samples are implemented to evaluate the robustness of MPSN. We also provide the mathematical solution of small perturbations to select more robust models. 

\textcolor{black}{Many studies have applied vision-based occupancy detection methods for occupancy-centric controls (OCC). Differently, this study provides new thinking to stabilize the detection and improve performance, filling the gap in video head detection at the pixel level. Through suppressing static background objects and enhancing moving occupants, MPSN is more robust and accurate, solving two essential problems (the false positive rate and false-negative rate) in building occupancy measurement to some extent. Besides, we try to open the black boxes of deep CNN and provide a detailed perturbation analysis in Appendix. Our experiments show a large potential in energy-saving and indoor environment comfort. A limitation of MPSN is that the time interval of adjacent frames is a constant. In the future, we would distill and deploy MPSN in edge devices (e.g., Raspberry Pi and Jetson Nano) in buildings. We hope our study can promote the application of the deep learning model in real-world building occupancy measurement. }

\textcolor{black}{
\section{ACKNOWLEDGMENTS}
This work was supported by the National Natural Science Foundation of China under Grant 62192751 and 61425027, in part by the National Key Research and Development Project of China under Grant 2017YFC0704100, and 2016YFB0901900, in part by the 111 International Collaboration Program of China under Grant BP2018006, in part by the BNRist Program under Grant No. BNR2019TD01009, and in part by China National Innovation Center of High Speed Train Project Under Grant CX/KJ-2020-0006.}

\textcolor{black}{We sincerely appreciate reviewers, whose valuable comments have benefited our paper significantly!}

\bibliographystyle{cas-model2-names}

\bibliography{cas-refs}

\clearpage
\appendix
\section{APPENDIX}

\subsection{Network architectures and hyperparameters}

Our detection network is described in Section~\ref{Method}. The VGGNet16, MobileNetv2, and Resnet18 backbone networks are applied for comparison. \textcolor{black}{They are pre-trained on ImageNet \citep{ILSVRC15} dataset.} Their detailed network architectures are shown in Table~\ref{threenets}. The architecture of $\emph{N}_{FDN}$ is the same as $\emph{N}_{FN}$'s, but the weights will be updated separately. If the resolution of the original image is $H\times W$, it will decrease to $H/4\times W/4$ after $\emph{N}_{FN}$ and $\emph{N}_{FDN}$. Similarly, the resolution will decrease to $H/16\times W/16$ after $\emph{N}_{BN}$ and $\emph{N}_{BDN}$. The entire model is trained by the SGD optimizer for 50 epochs. The learning rate is $10^{-2}$ and is decayed as $10^{-3}$, $10^{-4}$ and $10^{-5}$ after 15, 35 and 42 epochs separately. 
\textcolor{black}{The detector network contains 5 convolutional layers, which are initialized with standard normal distribution with 0.01 standard deviation.} MPSN is also tested on DETR. We select the Resnet18 as the backbone network, then add a pseudo-siamese network and DFA to \textcolor{black}{compare to the detection performance}. The model is trained for 300 epochs. The learning rate of the backbone network is set to $10^{-5}$ and is decayed as $10^{-6}$ and $10^{-7}$ after 100 and 200 epochs separately. The batchsize is set to 1.

\begin{table}[h]
\caption{Architectures of backbone network.}
\vspace{-1em}
\begin{center}
\begin{tabular*}{\tblwidth}{@{}LLLL@{}}
\toprule
Backbone&VGGNet16&MobileNetv2&Resnet18\\
\midrule
$\emph{N}_{FN}/\emph{N}_{FDN}$&\makecell[l]{conv1 \\ conv2 }&\makecell[l]{conv1 \\ bottleneck1-3 }& \makecell[l]{conv1 \\ conv2\_x}\\
\hline
$\emph{N}_{BN}/\emph{N}_{BDN}$&\makecell[l]{conv3 \\ conv4 \\conv5 }&bottleneck4-13 &\makecell[l]{conv3\_x \\ conv4\_x}\\
\bottomrule
\end{tabular*}
\vspace{-1.7em}
\label{threenets}
\end{center}
\end{table}

\subsection{\textcolor{black}{Gradient backpropagation}}
The derivatives of DFA in Section~\ref{Method} are related to the Hadamard product. \textcolor{black}{We} construct the diagonal matrix so that the following equation holds. 
\begin{equation*}
\bm{h_{bn}} \odot \bm{\sigma(h_{bdn})}={\rm Diag}(\bm{h_{bn}})\bm{\sigma(h_{bdn})}.
\end{equation*}
Besides, the element-wise function differential is 
\begin{equation*}
d(\bm{\sigma(h_{bdn})})=\bm{\sigma(h_{bdn})^{'}} \odot d(\bm{h_{bdn}}).
\end{equation*}
According to Eq.~\eqref{DFA}, we can obtain
\begin{equation}
\begin{aligned}
&\frac{\partial \bm{L(\theta,I_{f},y_f)}}{\partial \bm{h_{bdn}}} \\
&=\frac{\partial \bm{L(\theta,I_{f},y_f)}}{\partial \bm{h_{agg}}} \cdot \frac{\partial \bm{h_{agg}}}{\partial \bm{\sigma(h_{bdn})}} \cdot \frac{\partial \bm{\sigma(h_{bdn})}}{\partial \bm{h_{bdn}}} \\
&=\frac{\partial \bm{L(\theta,I_{f},y_f)}}{\partial \bm{h_{agg}}}\cdot [\alpha \cdot {\rm Diag}(\bm
{h_{bn}})+ \beta \cdot \bm{I}] \\
& \qquad \qquad \cdot {\rm Diag}[\bm{\sigma(h_{bdn})} \odot (\bm{1}-\bm{\sigma(h_{bdn})})],
\end{aligned}
\end{equation}
where $\bm{I}$ is the identity matrix, and $\bm{1}$ is the all-ones matrix.

\subsection{The mathematical solution of small $\epsilon$}
According to the Eq.~\eqref{motion} and Eq.~\eqref{attack input}, we have

\begin{equation*}
\begin{aligned}
&\bm{I_{df}^a}=|\bm{I_f^a}-\bm{I_{f-1}^a}|\\
&=|\bm{I_f}-\bm{I_{f-1}}+\epsilon [ {\rm sign}(\nabla_{\bm{I_f}} L(\bm{\theta},\bm{I_f},y_f))\\
&-{\rm sign}(\nabla_{\bm{I_{f-1}}} L(\bm{\theta},\bm{I_{f-1}},y_{f-1}))]|,
\end{aligned}
\end{equation*}
thus,

\begin{equation*}
\begin{aligned}
&\bm{I_{df}^a}-\bm{I_{df}}\leq|\bm{I_f}-\bm{I_{f-1}}| + \epsilon| {\rm sign}(\nabla_{\bm{I_f}} L(\bm{\theta},\bm{I_f},y_f))\\
&-{\rm sign}(\nabla_{\bm{I_{f-1}}} \!L(\bm{\theta},\!\bm{I_{f-1}},y_{f-1}))|-|\!\bm{I_f}-\!\bm{I_{f-1}}|\\
&=\!\epsilon| \!{\rm sign}(\nabla_{\bm{I_f}} L(\bm{\theta},\bm{I_f},y_f))- \!{\rm sign}(\nabla_{\bm{I_{f-1}}} L(\bm{\theta},\bm{I_{f-1}},y_{f-1}))|,
\end{aligned}
\end{equation*}

\begin{equation*}
\begin{aligned}
&\bm{I_{df}^a}-\bm{I_{df}} \geq |\bm{I_f}-\bm{I_{f-1}}| - \epsilon| {\rm sign}(\nabla_{\bm{I_f}} L(\bm{\theta},\bm{I_f},y_f))\\
&-{\rm sign}(\nabla_{\bm{I_{f-1}}} L(\bm{\theta},\bm{I_{f-1}},y_{f-1}))|-|\bm{I_f}-\bm{I_{f-1}}|\\
&=-\epsilon| {\rm sign}(\nabla_{\bm{I_f}} L(\bm{\theta},\bm{I_f},y_f))- {\rm sign}(\nabla_{\bm{I_{f-1}}} L(\bm{\theta},\bm{I_{f-1}},y_{f-1}))|.
\end{aligned}
\end{equation*}
Let $\bm{t}=| {\rm sign}(\nabla_{\bm{I_f}} L(\bm{\theta},\bm{I_f},y_f))- {\rm sign}(\nabla_{\bm{I_{f-1}}} L(\bm{\theta},\bm{I_{f-1}},y_{f-1}))|,$
thus,
\begin{equation}
-\epsilon\bm{t} \geq \bm{I_{df}^a}-\bm{I_{df}} \leq \epsilon\bm{t}.
\end{equation}

Because $\epsilon$ is small in applications, we conduct Taylor expansion at $\bm{I_f}$:

\begin{equation*}
\begin{aligned}
&g_\varphi(\bm{I_f^a})-g_\varphi(\bm{I_f})= \frac{\partial g_\varphi(\bm{I_f})}{\partial \bm{I_{f}}}^\top (\bm{I_{f}^a}-\bm{I_{f}})+\frac{\partial^2 g_\varphi(\bm{I_f})}{2 \partial \bm{I_{f}^2}}^\top (\bm{I_{f}^a}\\
&-\bm{I_{f}})^2 
+O(\epsilon^3)\approx \epsilon \frac{\partial g_\varphi(\bm{I_f})}{\partial \bm{I_{f}}}^\top {\rm sign}(\nabla_{\bm{I_f}} L(\bm{\varphi},\bm{I_f},y_{f-1}))+\\
&\frac{\epsilon^2}{2} {\rm sign}^\top(\nabla_{\bm{I_f}} L(\bm{\varphi},\bm{I_f},y_{f-1})) \frac{\partial^2 g_\varphi(\bm{I_f})}{2 \partial \bm{I_{f}^2}} {\rm sign}(\nabla_{\bm{I_f}} L(\bm{\varphi},\bm{I_f},y_{f-1})).
\end{aligned}
\end{equation*}
Similarly, we conduct Multiple Taylor expansion at $\bm{I_f},\bm{I_{df}}$:

\begin{equation*}
\begin{aligned}
&g_\theta(\bm{I_f^a,I_{df}^a})-g_\theta(\bm{I_f,I_{df}})=\frac{\partial g_\theta(\bm{I_f,I_{df}})}{\partial \bm{I_f}}^\top (\bm{I_f^a}
-\bm{I_f})+\\
&\frac{\partial g_\theta(\bm{I_f,I_{df}})}{\partial \bm{I_{df}}}^\top (\bm{I_{df}^a}-\bm{I_{df}})
+\frac{1}{2}[(\bm{I_f^a}-\bm{I_f})^\top (\bm{I_{df}^a}-\bm{I_{df}})^\top] \\ &
\begin{bmatrix}
\frac{\partial^2 g_\theta(\bm{I_f,I_{df}})}{\partial \bm{I_f^2}} &\frac{\partial^2 g_\theta(\bm{I_f,I_{df}})}{\partial \bm{I_f} \partial \bm{I_{df}}}\\
\frac{\partial^2 g_\theta(\bm{I_f,I_{df}})}{\partial \bm{I_{df}} \partial \bm{I_f}} &\frac{\partial^2 g_\theta(\bm{I_f,I_{df}})}{\partial \bm{I_{df}^2}}
\end{bmatrix}
\begin{bmatrix}
(\bm{I_f^a}-\bm{I_f})\\
(\bm{I_{df}^a}-\bm{I_{df}})
\end{bmatrix}
+O(\epsilon^3)\\
& \leq \epsilon \frac{\partial g_\theta(\bm{I_f,I_{df}})}{\partial \bm{I_f}}^\top {\rm sign}(\nabla_{\bm{I_f}} L(\bm{\theta},\bm{I_f},y_f))+\epsilon |\frac{\partial g_\theta(\bm{I_f,I_{df}})}{\partial \bm{I_{df}}}^\top| \\
& \bm{t}+\frac{\epsilon^2}{2} {\rm sign}^\top(\nabla_{\bm{I_f}} L(\bm{\theta},\bm{I_f},y_f)) \frac{\partial^2 g_\theta(\bm{I_f,I_{df}})}{\partial \bm{I_f^2}} {\rm sign}(\nabla_{\bm{I_f}} L(\bm{\theta},\\
&\bm{I_f}, y_f))+\epsilon^2 |{\rm sign}^\top(\nabla_{\bm{I_f}} L(\bm{\theta},\bm{I_f},y_f)) \frac{\partial^2 g_\theta(\bm{I_f,I_{df}})}{\partial \bm{I_f} \partial \bm{I_{df}}}| \bm{t}\\
& +\frac{\epsilon^2}{2} \bm{t}^\top \frac{\partial^2 g_\theta(\bm{I_f,I_{df}})}{\partial \bm{I_{df}^2}} \bm{t}.
\end{aligned}
\end{equation*}

When $\bm{{NI\_MPSN}} \leq \bm{{NI\_FCHD}} $:
\begin{equation}
\begin{aligned}
& \left\|\frac{1}{N C} \sum_{\bm{f}=1}^{N} [g_\theta(\bm{I_f^a,I_{df}^a})-g_\theta(\bm{I_f,I_{df}})] \right\|_2 \\ & \qquad \leq \left\|\frac{1}{(N+1) D} \sum_{\bm{f}=0}^{N} [g_\varphi(\bm{I_f^a}) -g_\varphi(\bm{I_f})] \right\|_2.
\end{aligned}
\end{equation}
Both sides are quadratic functions of $\epsilon$, thus,

\begin{align} 
|j\epsilon + \epsilon^2 k| \le |l \epsilon+ \epsilon^2 m| \Rightarrow |j+ \epsilon k| \le |l + \epsilon m|.
\label{inter}
\end{align}
The small $\epsilon$ can be obtained mathematically.

\bio{}

\endbio

\end{document}